%% file: extended_abstract.tex
\documentclass[runningheads]{llncs}
\usepackage{graphicx}
\usepackage{tikz}
\usepackage{url}
\usepackage{eurosym}
\usepackage{fancyvrb}
\usepackage{listings}
\usepackage{graphicx}
\usepackage{enumitem}
\usepackage{amsmath}
\usepackage{amssymb}
\usepackage{listings}
\usetikzlibrary{shapes,arrows}
\setlist{itemsep=0pt,parsep=0pt,topsep=.5\topsep,leftmargin=\parindent}
\usepackage{blindtext}
\begin{document}
\input{macro}
%
\title{Design Space Exploration via Answer Set Programming Modulo Theories}
\author{Philipp Wanko}

\institute{Institute of Computer Science, August-Bebel-Str. 89, 14482 Potsdam, Germany}

\maketitle              % typeset the header of the contribution

\begin{abstract}
	The design of embedded systems, that are ubiquitously used in mobile devices and cars,
	is becoming continuously more complex such that efficient system-level design methods are becoming crucial. 
	My research aims at developing systems that help the designer express the complex design problem in a declarative way 
	and explore the design space to obtain divers sets of solutions with desirable properties.
	To that end, we employ knowledge representation and reasoning capabilities of ASP in combination with background theories.
	As a result, for the first time, we proposed a sophisticated methodology that allows for the direct integration of multi-objective optimization of non-linear objectives into ASP.
        This includes unique results of diverse sub-problems covered in several publications which I will present in this work.
\end{abstract}

\section{Introduction}

With increasing demands for functionality, performance, and energy consumption in both industrial and private environments,
the development of corresponding embedded processing systems (ECSs), that are for example used in mobile devices or cars, is becoming more and more intricate.
Also, desired properties are conflicting and compromises have to be found
from a vast number of options to decide the most viable design alternatives.
Hence, effective Design Space Exploration (DSE;~\cite{pimentel17a}) is imperative
to create modern embedded systems with desirable properties;
it aims at finding a representative set of optimal valid solutions to a design problem
helping the designer to identify the best possible options.

The overall aim of my research until now was the study of exact DSE techniques for ECSs
utilizing the knowledge representation and reasoning capabilities of ASP
in combination with background theories (\emph{ASP modulo Theories}, ASPmT).
As a result, for the first time, we proposed a sophisticated methodology
that allows for the direct integration of multi-objective optimization of non-linear objectives into ASP.
This includes unique results of diverse sub-problems covered in several publications.

\section{ASPmT-based System Synthesis}
Our first objective was the development of an ASPmT-based approach to system synthesis
including allocation, binding, routing, and scheduling while adhering to area, energy, and timing constraints.
Allocation determines the hardware components that are used from a given architecture,
binding assigns applications to hardware components,
routing determines the paths of messages between applications through the hardware architecture,
and finally, scheduling determines the exact time points when applications are executed and messages sent.
Area, energy, and timing constraints refer to the minimization of the number of hardware components,
energy consumption, and duration of the execution of the applications, respectively.
Our results, published in \cite{newascha17a}.
We holistically encode the system synthesis problem featuring meshed network-on-chip (NoC) hardware architecture and multi-hop communication,
as well as conflict-free scheduling of periodically activated applications.
The schedulability analysis is executed as two background theories,
specifically by a propagator handling quantifier-free integer difference logic (QF-IDL) constraints and another checking periodic overlaps.
The scheduling is tightly integrated and capable of partial solution checking by using our ASPmT framework~\cite{gekakaosscwa16a};
it extends the ASP system clingo with a general interface to integrate application- and theory-specific reasoning into ASP.
That is, arbitrary theories can now be directly included into ASP's modeling language through common couple variables
that are part of standard ASP rules and thus,
can profit from the sophisticated propagation techniques offered by clingo.

We delegate aspects of the system synthesis problem to standard ASP,
namely binding and routing as well as area and static energy requirements,
since they can be effectively expressed and verified in pure ASP (cf.\ \cite{angeglharesc13a})
while we use QF-IDL for timing constraints.
Thus, we are alleviating the need to express all possible start times directly in ASP, which would lead to a blowup in problem size,
and assigning non-linear scheduling tasks to a more suitable paradigm.
QF-IDL furthermore has the advantage of being solvable in polynomial time and the consistency checking yields minimal conflict clauses which are imperative for solution space pruning.

In fact, our contribution goes well beyond DSE:
The system clingo[DL]~\cite{jakaosscscwa17a} instantiates the ASPmT framework of clingo with QF-IDL,
thus providing a generic combination of ASP with QF-IDL, featuring a hybrid modeling language along with hybrid multi-objective optimization.
Moreover, our experimental studies showed that clingo[DL] performs very well compared to other state-of-the-art hybrid solvers,
including other instantiations of clingo's ASPmT framework,
using linear programming and constraint programming for expressing difference constraints,
as well as comparable state-of-the-art hybrid solver based on modern SMT-solvers.

\section{ASPmT-based Design Space Exploration Framework}

After studying ASP-based system synthesis,
we extended our approach towards a holistic DSE framework including multi-objective optimization.
The results have been presented in \cite{newascha18b}.

I will now outline alternative methods for computing Pareto-optimal solutions in our DSE framework.
Ideally, after the design space is explored completely,
DSE returns the true Pareto set containing all optimal solutions
from which the designer can choose the favored design points.
However, an exhaustive search in the design space is not viable for large problem instances,
which is why only an \emph{approximate} Pareto set is obtainable in reasonable time.
Hence, we developed three different exploration strategies calculating the Pareto set in an anytime fashion.
That is, whenever DSE is aborted prematurely, an approximate Pareto set is returned;
its quality improves over time until eventually the true Pareto set is found.
We evaluate all alternatives by two criteria:
entropy \cite{fa:2003} specifying how regular the design points are distributed across the objective space (diversity),
and $\epsilon$-dominance \cite{ztlfg:2003} reflecting the distance between true Pareto and approximation set (convergence).

The first strategy is based on asprin~\cite{brderosc15a}, a general framework for optimization in ASP.
In asprin, ASP is used to express objectives, and optimization follows a branch and bound scheme.
That is, each new solution is required to be better than the previous one, until no better one is found.
We extend this approach by allowing for propagators to implement objectives that depend on other background theories.
This is needed, e.g.,
for optimizing latency since the starting times of tasks and communications are only known to the QF-IDL theory.
Regardless of the origin of the preference,
it is used to derive special atoms whenever the current solution improves a previous one.
These atoms are in turn aggregated by an ASP rule to express Pareto preference.
Consecutively, constraints are added to the problem definition
ensuring an improvement of a solution, until a true Pareto-optimal design point is found.
This approach first tries to prove Pareto optimality by improving convergence continuously.
To enumerate Pareto optimal solutions,
a constraint is added whenever an optimal solution is found,
enforcing that newly found solutions are incomparable to the optimum.
After that, the branch and bound process resumes until no more solutions are found.
% Previous solutions are partly saved as facts within the logic program rendering an explicit archive in the background theory obsolete.
%

The second approach follows the same optimization scheme
% i.e., it only searches for incomparable solutions whenever a true Pareto-point is found.
% In contrast to the previous approach,
but evaluates the objectives as well as the dominance checks within the background theory.
The third approach explores the design space in a breadth-first fashion,
i.e., if a new design point is found, it is not required to strictly dominate previously found solutions.
All currently known best solutions are kept in an archive
and dominated solutions are removed whenever a better solution is found.
Similar to the second approach,
evaluation of design points and enforcement of quality constraints are exclusively handled in the background theory.

Our experiments show that the third strategy finds the most diverse solutions~\cite{newascha18b}.
Considering convergence, the first and third approach perform nearly the same.
However, the former only returns one best known solution after the timeout
while the latter offers an approximate Pareto set with multiple solutions.
% Therefore, the breadth-first approach is considered to be superior.
The experiments show that we are able to handle large problem instances with up to 170 tasks
mapped to a hardware platform implemented as a meshed NoC as communication infrastructure.

Compared to existing meta-heuristic \cite{btt:1998}, hybrid \cite{lght:2007a,slht:2006,nhg:2016} and exact \cite{lght:2008} DSE techniques,
our approach has several advantages.
First, even for highly constrained problems, found solutions are guaranteed to be feasible.
Second, design points are not explored multiple times resulting in a more efficient exploration.
Third, reachability can be directly expressed in ASP.
That is, the encoding for routing of communication messages is formulated naturally with recursive rules.
Finally, since ASPmT allows us to conduct constraint and dominance checks on partial assignments,
our approach prunes larger regions earlier during search.

As the number of non-dominated design points increases during search,
archiving them becomes tedious due to the growing number of dominance checks.
This is even more severe when dealing with partial assignments since each undergoes a dominance check until a complete solution is obtained.
We address this in \cite{nehawasc18a} by investigating \emph{Quad-Trees} as basic data structure of solution archives.
Our experiments show that they help diminishing the number of comparison operations by an order of magnitude
when considering more than three objective functions.
In contrast to using Quad-Trees with complete solutions only, as done in \cite{Mostaghim2005},
updating the archive of solutions does not have to be performed for each dominance check.
The reason for this is that a partial solution may dominate the archive but deteriorates with additional decisions made.
Thus, it cannot displace any solutions from the archive until all decisions are made.
That is, the dominance check of partial solutions only signifies the necessity of assigning the remaining decisions.
This allows for skipping the expensive update operation of the archive for partial assignments and executing it only in the final step.

\section{Related Publications}

To warrant the generality of our techniques, we elaborated upon Curriculum-Based Course Timetabling (CB-CTT) in \cite{bainkaokscsotawa18a},
a problem closely related to schedulability analysis. % in embedded systems design.
% This resulted in the system \emph{teaspoon} takes a standard CB-CTT format as input and solves the problem using ASP.
While the hard constraints are fixed,
the inclusion of soft constraints and their priority can be freely configured.
% At its core, the problem tries to assign lectures of courses to time slots and rooms while avoiding scheduling conflicts,
% e.g., no room is assigned twice at the same time and the teacher of the course is available.
% Additionally, courses are grouped in curricula and different courses of a curriculum may not be scheduled simultaneously.
% The soft constraints pertain to the suitability of rooms for certain courses,
% the optimal spacing of the lectures in the schedule, and the workload of the students.
We were able to draw from a well of existing real-world benchmarks stemming from the CB-CTT community,
for which best known schedules have already been obtained by a plethora of different technologies.
Besides improving or reproducing best known bounds for over half of the available instance set,
our system solved difficult combinations of soft constraints for very large instances,
like the benchmarks from the University of Erlangen, for the first time.
We were also able to improve solutions for instances with unknown optima by using approximate optimization schemes.
In detail, instead of aggregating all soft constraints, we ordered them lexicographically.
While the optima of the lexicographic optimization might not provide the globally best bound,
by separating the objective functions and optimizing difficult criteria last,
we improved the best known bounds for large instances.

Often, a valid schedule is already in place but circumstances change.
% e.g., teachers or rooms are no longer available, courses are added or removed,
% or further soft constraints are added to improve schedule quality.
In such a case, it is desirable to change the previous schedule as little as possible (stability),
while finding a schedule fulfilling the new constraints with high quality (optimality).
This problem is called \emph{Minimal Perturbation Problem for Curriculum-Based Course Timetabling}. % (MPP CB-CTT).
With our ASP techniques,
we were able to implement a solving scheme based on lexicographic optimization that enumerates all Pareto optimal solutions regarding stability and optimality.
The idea is to first obtain the extreme Pareto optimal solutions
by calculating the lexicographic optima.
Then, the bounds of these solutions are used to restrict the search space,
cutting off dominated areas and lexicographic optimization resumes.
The full Pareto set is found once no solution is found in the restricted search space.

To present small subsets of representative solutions,
we introduce a framework for computing diverse (or similar) solutions to logic programs with preferences in \cite{roscwa16a}.
The first contribution is the automation of various ASP solving schemes,
namely max-min optimization, guess and check, querying and preferences over preferences.
Using the solving schemes as building blocks,
the framework provides three kinds of techniques:
enumeration, replication and approximation.
Each one tries to calculate $n$ most diverse/similar optimal solutions
given a logic program with preferences and a distance measure.
The latter gives the pair-wise distance between two solutions,
e.g., the number of varying truth values in two solutions.
A set of solutions is most diverse when the minimal pair-wise distance
is maximal among all possible solution sets of cardinality $n$.
Our experiments show that enumeration and replication are ineffective,
since exponentially many optimal solutions might have to be enumerated,
and treating series of already complex optimization problem leads to an explosion in complexity.
Approximation was the most successful technique,
and we witnessed a trade-off between diversification quality and run time
depending on whether optimization or heuristics were used to identify the next optimal solution.

To evaluate our approaches,
it is imperative to have a significant amount of test cases that allow for studying performance and scalability.
We therefore investigated a methodology to systematically create system specification instances
that results in a versatile and easily expandable  ASP-based benchmark generator~\cite{nehawasc18c}.
The generator produces synthetic system synthesis specifications
composed of applications, heterogeneous hardware platforms, and mapping options.
Due to its modular structure, rules for encoding the generation of applications or hardware platforms can be adjusted independently of each other,
which allows us to utilize the framework for a wide range of application and hardware platform structures.
By default, it generates applications with series-parallel communication patterns
that allow for modeling a plethora of characteristics with one versatile ASP encoding.
Depending on the dominance of series or parallel patterns,
the application allows for more or less concurrency.
Utilizing ASP, the generator produces variant instances with similar characteristics (e.g., number of tasks, connectivity, concurrency)
while the shapes of the generated applications differ.
To check produced instances w.r.t.~constraints such as minimum latency,
we combined the generator with the ASPmT-based system synthesis approach.

In order to improve DSE,
we currently investigate over and under estimation techniques for quality constraints and objective functions.
The rational behind such estimation techniques is a more rapid evaluation phase of a found design point.
Additionally, if an estimation is guaranteed to evaluate an objective as better compared to the exact value,
e.g., under estimation for minimization problems,
we use this information to skip expensive calculations without the loss of accuracy of the obtained Pareto optimality \cite{nwsh:2018}.
This is particularly true for partial assignments that have to be evaluated many times before the complete solution is obtained.
That is, if the under estimation evaluation is already dominated by a design point in the archive,
the exact evaluation would be worse and thus also be dominated.
Hence, the expensive exact calculation can be skipped.
Only if the under estimation indicates a good solution, the exact calculation has to be performed.

\section{Future Work}

My current goal is developing a theoretic framework based on the Logic of Here-and-There~\cite{heyting30a,pearce96a,pearce06a} capable of capturing the combination of arbitrary theories and sophisticated language constructs.
This should enable, first, a foundation for seamless combination of multiple theories, e.g., QF-IDL with ILP, and second, advanced language constructs like aggregates involving entities related to the theories, e.g., an aggregate calculating the maximum over variables used in difference constraints.
Eventually, this will be realized  via clingo's theory API and function as a culmination and combination of my doctoral studies.

%%% Local Variables:
%%% mode: latex
%%% TeX-master: "proposal"
%%% End:

\input{extended_abstract.bbl}
\end{document}

%% file: macro.tex
\newcommand{\workarea}[2]{\subsubsection*{#1.\ #2}}
\newcommand{\workpackage}[3]{\paragraph*{#1.\ #2\hfill\textnormal{(#3)}}}
\newcommand{\ExpectedResults}{\par\noindent\textbf{Results:}} % {\par\noindent\textbf{Expected Results:}}
\newcommand{\ProposedApproach}{\par\noindent\textbf{Approach:}} % {\par\noindent\textbf{Proposed Approach:}}
\newcommand{\ProblemStatement}{\par\noindent\textbf{Problem:}} % {\par\noindent\textbf{Problem Statement:}}
\newcommand{\myparagraph}[1]{\par\smallskip\noindent\textbf{#1}}

%%% Local Variables:
%%% mode: latex
%%% TeX-master: "proposal"
%%% End: